\documentclass[letterpaper]{article}
\usepackage{natbib,alifexii}   %% The order is important
\pdfoutput=1
\title{Polyethism in a colony of artificial ants}
\author {Chris Marriott \and Carlos Gershenson\\
\mbox{} \\
IIMAS, Universidad Nacional Aut\'onoma de M\'exico, M\'exico City, M\'exico \\
algorithm0r@gmail.com, cgg@unam.mx}

\begin{document}
\maketitle

\begin{abstract}
We explore self-organizing strategies for role assignment in a foraging task carried out by a colony of artificial agents.  Our strategies are inspired by various mechanisms of division of labor (polyethism) observed in eusocial insects like ants, termites, or bees.  Specifically we instantiate models of \emph{caste polyethism} and \emph{age} or \emph{temporal polyethism} to evaluated the benefits to foraging in a dynamic environment.  Our experiment is directly related to the exploration/exploitation trade of in machine learning.
\end{abstract}

\section{Introduction}

The self-organizing strategies of eusocial insects are now well known and well studied in biology (\cite{BGDP89,T89,R92,TBD98,TB99,GTDA02,GGT07}) and applications to computation are abundant (\cite{BDT99,PL04a,PL04b,SC08,G10,DDG10}).  One of the more remarkable behaviors observed is the ability of rather simple, unintelligent agents (individual insects) to coordinate their behavior to establish a rather fluid and adaptive behavior on the colony level.  The phenomenon of \emph{stigmergy} (communication via the environment) has now been modeled and applied in artificial simulations to achieve similar results among rather simple artificial agents (\cite{TB99,BDT99,PL04a,PL04b,SC08}) cooperating in multi-agent systems.

However, many of these applications focus on homogeneous colonies, where each agent has the same behavioral capabilities.  Nonetheless, observations of insects show that in many colonies the individuals are not always homogeneous.  Colonies consist of heterogeneous agents, whether these agents display morphological differences (i.e. distinct castes) or merely behavioral differences.  The effects of this stratification of agents in a colony is referred to as division of labor (DOL) or by the term \emph{polyethism} (\cite{R92,TR97,TBD98,GTDA02,G03}).  As artificial multi-agents systems grow larger and involve agents with different roles the problem of assigning roles to agents becomes increasingly important (\cite{CW10, DB09}).

Biologists differentiate between at least two observable means of dividing roles amongst individual workers in natural insect colonies.  The means we select for study are called \emph{caste polyethism} and \emph{age polyethism}.  Other types of polytheism are also observed (e.g.\,\emph{elitism}) and the two above types have many possible underlying mechanisms though these additional types and subtypes will not be explored in detail in this article.  Simulations have just begun exploring task assignment and heterogeneous agent populations (e.g.\,\cite{SC08,DDG10}).  Our experiment differs from these in that our agents are assigned the same task (foraging), but must decide which strategy  to adopt to solve the task(between an individual exploratory strategy and a cooperative exploitative strategy).  In this sense, our experiment parallels attempts to solve the well known exploration/exploitation trade off in machine learning.  Further, other experiments focus on simulations of actual natural colony behavior in an attempt to assess models of those behaviors, whereas while we are inspired by these models our focus is on the self-organizing and adaptive problem solving that these models make possible.

\emph{Caste polyethism} occurs when distinct types of individuals are bred by the colony.  An individual is effectively born into its role, often times displaying morphological differences from individuals from other castes.  The clearest example of castes is the division between the reproductive caste and the worker caste in eusocial insects.  A single or small group of reproductive females (called queens) are responsible for all reproductive tasks in the colony while non-reproductive workers carry out all other tasks required by the colony (brood care, nest constructions and maintenance, waste removal, foraging, and defense).  In some species workers are further divided into sub-castes.  Differences among workers from different castes are particular to the worker's role.  For instance in some species of ants the workers can be divided into \emph{majors} and \emph{minors} (occasionally with an intermediate caste as well) where the majors are larger than the minors, this size being helpful in the task they carry out (primarily colony defense).  Minors are smaller, making them more energy efficient, and they are relegated to less dangerous tasks like foraging and nest maintenance.  Only in rare occasions will a worker do a task that is typically assigned to a different caste.

\emph{Age} or \emph{temporal polyethism} is a type of division of labor where the worker's role is correlated with its age or changes over time.  Age polyethism is more common than caste polyethism in natural insect colonies.  In colonies displaying age polyethism younger workers are commonly assigned less risky tasks (nursing or nest maintenance allowing them to stay in the nest) whereas older workers are assigned more dangerous tasks (foraging, defense, or raiding where the agent must leave the nest).  It is hypothesized that this division of labor allows the colony to maximize the work carried out by each individual worker (i.e. young workers will be less likely to die and thus can live longer to carry out more work).  This will be beneficial to the colony since it will have to breed fewer workers if each worker's longevity (and thus productivity) is maximized (\cite{T02,T09}).  In certain cases this progressive role assignment may also allow younger and less experienced workers to gain the experience necessary to carry out more difficult tasks (say at the very least allowing them to become familiar with the layout of the nest and surrounding environment before having to venture far from the nest) (\cite{TF92,T93,FT94}).  Many mechanisms have been suggested as the underlying reason for observed age polyethism.  The mechanism we employ is similar to the response threshold model commonly studied (see e.g. \cite{TBD98,GGT07}).

As systems of artificial agents grow larger (in population) and become more heterogeneous the task of assigning roles to agents becomes more critical.  This article aims to explore models that might achieve the types of division of labor observed in eusocial insects so that these models may be exploited in engineering of multi-robot and multi-agent systems.

\section{Artificial Ants}
The experiment detailed below involves a colony of artificial ants engaged in a foraging task.  The colony level task is to maximize the food intake of the colony (allowing colony sustenance and growth).  On the individual worker level the task is to explore the environment, find a food object, and return to the nest with the object.

We consider two different strategies for individual workers inspired by natural ant populations.  The first, and simpler, strategy is for workers to forage for the most part individually.  We say ``for the most part'' here since individual foragers cooperate at least insofar as they attempt to divide the environment to be explored equally among them (see Figure 1).  We implement this strategy by having ants leave a ``seeker'' trail as they leave the nest.  While ``seeking'' the ants will avoid other seeker trails, meaning they will travel mostly straight away from the nest while avoiding the trail they leave behind them, but they will also avoid trails left by other ants, helping to divide the area somewhat evenly.  Other than this simple cooperation, workers leave the nest and randomly explore until they find a food object (or reach the range of their exploration) and return to the nest.  We will call this strategy the ``individual'' or ``exploratory'' strategy, and ants following this strategy ``explorers''.  The seeker path left by these ants also serves as the ants sole means of returning to the den (i.e. they follow seeker paths back).

Ants that find a food source of sufficient size (i.e. they find at least one food morsel to carry back to the nest and at least one more food morsel they will recruit others to seek out) the ant will leave a second type of trail we call the ``carrier'' trail.  The second strategy, which we call the ``cooperative'' strategy or ``exploitative'' strategy, involves foragers that will follow ``carrier'' trails to exploit food sources that were already discovered by other ants.  Both explorers and exploiters will leave ``carrier'' trails under the conditions listed above, but only exploiters will follow them to food sources (explorers ignore them).  

These strategies are inspired by those found in natural populations, with a correlation of colony size to the strategy used (\cite{BGDP89, T89}).  In particular it has been observed that smaller colonies tend to use the individual exploratory strategy whereas as the larger the colony is the more likely the colony uses a cooperative or exploitative strategy (and the more cooperative the strategy used).  Despite this correlation, upon closer examination larger colonies have foragers carrying out both strategies, that is, they engage in DOL or polyethism.

\begin{figure}
\begin{center}
\includegraphics[width = 3in]{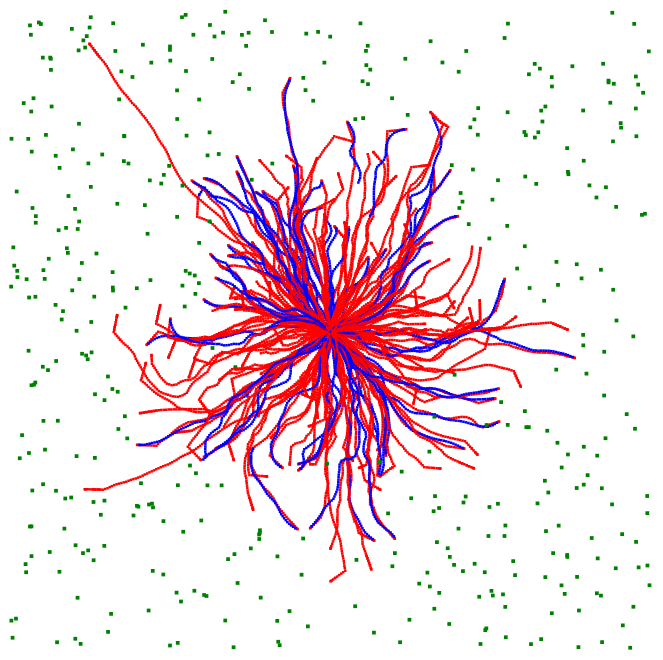}
\caption{Explorers in a uniform environment.  The den is in the center of the torus.  Green squares are food.  Red paths are seeker paths.  Blue paths are carrier paths.  Recall that explorers ignore the carrier paths.}
\end{center}
\end{figure}

In preliminary experimentation it was found that these strategies fare differently depending on the environment the colony is situated in.  If food objects are uniformly distributed around the nest then the individual strategy reaches near optimal foraging.  Over time the workers will clear a disc shaped area of food around the nest, the radius of the disc being determined by the frequency of food objects and by the size of the population.  This situation is presented in Figure 1.

Interestingly, in larger colony sizes the cooperative strategy also fares quite well in environments with uniform distribution of food, though the foragers carry out a more complex foraging strategy.  Cooperative foragers form an ``arm'' leading from the nest into the environment and this arm has been observed to swing in a circle around the nest, clearing food objects as it goes, or spontaneously dissolving and reforming in a more lucrative direction.  These strategies have also been observed in natural ant colonies.  While the cooperative strategy seems to approach the performance of the individual strategy in experimentation, the individual foragers have an advantage in an environment with uniformly distributed food.

\begin{figure}
\begin{center}
\includegraphics[width = 3in]{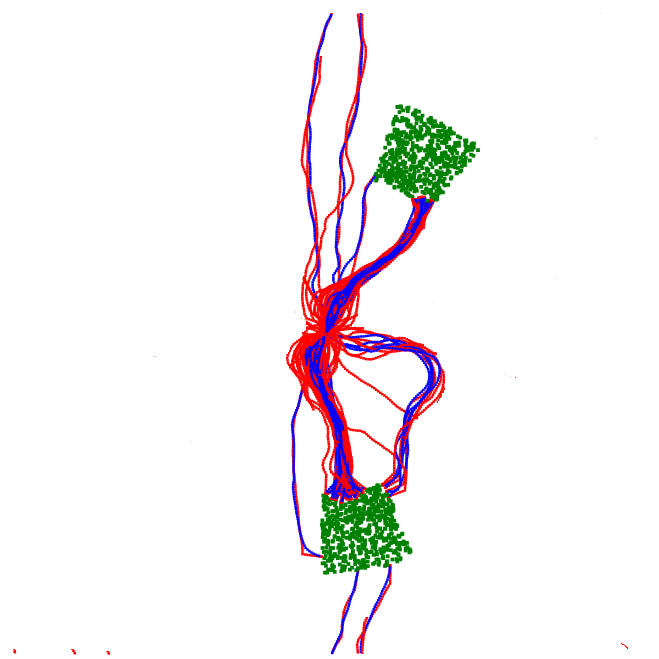}
\caption{Exploiters in an environment with two patches.  The den is in the center of the torus.  Green squares are food.  Red paths are seeker paths.  Blue paths are carrier paths.  Exploiters use the carrier paths to cooperatively forage.}
\end{center}
\end{figure}

A second environment type we have investigated contains food isolated in ``patches''.  For the sake of comparison among simulation runs our food patches are always placed equidistant from the nest, though in a random direction.  In this environment the cooperative foragers have a clear advantage.  Once a forager finds a patch of food it recruits other foragers to help it clear the patch and the colony quickly optimizes the path to the food patch.  Figure 2 shows a typical patch environment (with 2 patches) and a colony of exploitative ants foraging from the patches.

Individual foragers are at a significant disadvantage when faced with an environment with a single patch.  Many individual foragers leave the nest in the wrong direction (remember they attempt to divide the environment equally) and so return empty handed.  Only a fraction of individual foragers leave the nest in the right direction and return with food.

Given the differential success of these strategies in these environments it is our hypothesis that polyethism in a colony will be beneficial if the colony is faced with either an unknown environment (of one of these two types) or with a dynamic environment consisting of either a combination of these types or shifting between these types.

\section{Experimental Setup}

In our experiment we consider four different types of colonies that we will expose to five different types of environment.  We will consider how each colony fares in each environment, as well as how the colony fares across all environments.

A colony will consist of a queen (responsible for creating new workers), a population of workers, a population of larvae, and a store of food.  Workers consume food at a constant rate (about 1 food every 450 simulation rounds) and larvae consume food at a constant rate (1 food for the 100 round gestation period) until they are born as a new worker.

The queen lives for the duration of the experiment (or until the colony dies of starvation), though workers and larvae may die.  Workers die under two conditions.  If they reach their maximum age (selected uniformly from the range 2750-3250 rounds), or if they run out of food energy.  When a worker consumes a piece of food it gains energy that will sustain it for 450 simulation rounds.  If while foraging the worker's food energy reaches 0 (i.e. after 450 rounds) then the worker attempts to return to the nest (possibly without food).  Upon returning the worker will attempt to consume a unit of food from the store.  If there is no food in the store the worker dies.

A larvae also consumes food, once upon creation by the queen and again upon changing into a worker.  The food consumed when the larvae matures forms the initial energy store of the worker.  A queen will never create a larvae in an instance where the food stores are empty, however, a larvae may mature and find the store empty.  In this case the new worker dies.

Queens from different colonies have different profiles, however, they all follow the same rule when deciding to reproduce.  A queen will only create a new larvae if the food store exceeds the current population of workers plus the current population of larvae.

The first two types of colony will form a control group for comparison.  These two types will not use polyethism and queens in these colonies will create only explorers or only exploiters respectively.  From the earlier discussion we know that these colonies will fare well in some environments but not in others and will not be adaptive to the environment.

The third colony will engage in an adaptive caste polyethism.  Queens in this type of colony produce larvae that can mature into either an individual or cooperative worker.  The queen chooses the type of worker to create in proportion to the success rate of workers of that type.  (The queen keeps track of food returned by each type of forager over the last 500 rounds, and of the number of each type of forager.  From this she estimates the efficiency of the average ant of each type and randomly selects to create a new ant in proportion to the ratio of success rate.) Thus if explorers are more successful at foraging than exploiters then a queen will make an explorer with higher probability (and vice versa).  Queens in this type of colony will ensure there is always at least one worker of each type so success rates can be properly estimated. 

The fourth colony will engage in one type of age polyethism.  Workers in these colonies are homogeneous in their behavioral repertoire, in that they can act as either explorers or exploiters.  Which role a worker adopts depends first on their age (for younger workers) and then on the demands of the colony (for older workers).  In this colony new workers adopt an individual foraging strategy, and may switch to a cooperative strategy (or back again) after reaching a particular age (usually consisting of 1 or 2 full foraging trips).  Workers of this type choose to change roles based on collective experience, that is, in proportion to the success rate of workers in the colony similar to the mechanism used in the third colony.  While we do not use pheromones in our model of this behavior we believe this mechanism is closely related to response threshold models of behavior selection.

We expose these 4 colony types to 5 distinct environments: uniform, patch, roaming patch, seasonal, and mixed.  The rate at which food drops in each environment is the same (1 food every 5 rounds) and each food will stay in the environment for exactly 1000 rounds or until picked up by a forager.  The uniform and patch environments were described above consisting of uniformly distributed food or an isolated patch of food respectively.

The roaming patch environment as has a single patch but this patch will change location every 1000 rounds (the new location will be the same distance from the nest as the old location).  This means that after the patch has moved new food will drop in the new patch location, though old food is not removed unless foraged or it reaches its 1000 round limit.  As a result there will usually be two patches in the environment, one containing old food that is decaying and one containing new food.  Figure 2 displays a typical scenario for this type of environment.

The seasonal environment is intended to simulate an environment that changes from a uniform distribution to an isolated patch with regularity possibly corresponding to the seasons.  We simulate this idea by alternating between the two distributions every 1000 rounds.  Again there will be a temporal overlap between these two environments meaning that the environment will typically contain food dropped uniformly and in a patch.  Every time the season changes to the patch distribution a new location for the patch is selected so in this sense we see the patch as roaming as in the last environment.

The mixed environment includes both uniform food drops and an isolated patch at the same time, and the environment is static (in that the patch does not move).  In this environment the drop rate is the same as previous environments despite there being two active food drop mechanisms operating simultaneously.

\section{Observations and Data}

We choose to analyze the worker population data of our colonies.  This data reflects the colonies' ability to forage for food efficiently.  Each colony begins with an initial food store of 32 food and zero ants.  The queen will use this initial food to create 16 initial workers which mature on rounds 101-116 of the simulation.  At this point the food stores will be exhausted (each larvae will use one food when created and another when maturing into a worker) and so the colony must forage for food to sustain itself or grow in size.  Also by round 116, 23 food objects will exist in the environment, with their location depending on the type of environment.  Parameters of the simulation determine a maximum colony size, namely the food drop rate and the energy consumption rate of the workers (as well to a lesser extent the size of the environment).  This maximum is just above 80 workers, though due to the non-linear dynamics of the simulation this maximum can be exceeded for short periods.

The initial stages of the simulation are occupied by rapid growth of population as the foragers are able to bring in more food than the colony needs so new workers are created (exceptions to this are noted below).  This rapid growth commonly results in too many workers and so is often followed by a large dip in population and an oscillation is observed until an equilibrium can be found.  This equilibrium depends on the type of colony and environment.

\begin{figure*}
\begin{center}
\includegraphics{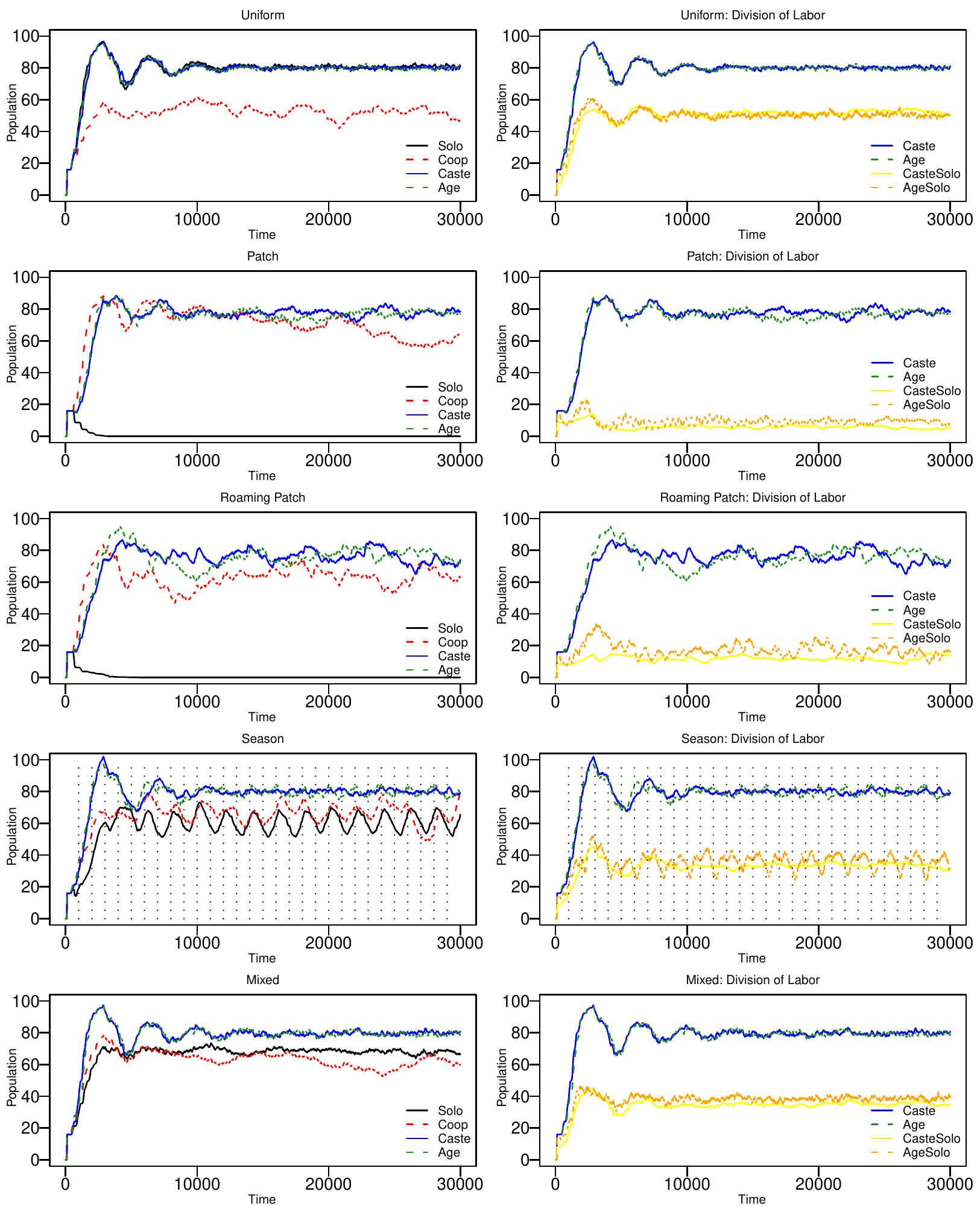}
\caption{Worker Population Data.  From the top row the data is presented for each environment: uniform, patch, roaming patch, seasonal, and mixed.  The left column displays worker population over time for the four colony types.  The right column displays the division of labor in the Caste and Age colonies.  The worker population of these colonies is contrasted to the number of workers in the colony assigned to the exploration task.  Dotted lines in the seasonal environment indicate the changing seasons.  Please note we use "Solo" to indicate explorers and "Coop" to indicate exploiters in the charts.}
\end{center}
\end{figure*}

Figure 3 (left column) displays the worker population data gathered from all experimental runs.  The data presented in the figure is the average worker population over time ($N = 13$).

Beginning with our control environments, we note that the colonies perform as expected.  In the uniform environment the best performance is achieved by the explorers, and is closely matched by the caste and age polyethistic colonies.  All three colonies settle around a population of 80 workers after initial instability.  While the exploitative colony has no trouble surviving in this environment its sub-optimal foraging strategy allows it to maintain only a population of between 40-60 workers.  It's population is also subject to greater instability as the foraging arm grows and shrinks in size and changes location.

In the second control environment with a single stationary patch again we see expected results.  The explorers are unable to maintain even the low initial colony size and the colony starves quickly.  The cooperative foragers are the quickest to exploit the isolated patch, whereas the polyethistic colonies are able to quickly adapt to the environment by producing exploiters instead of explorers.  Both polyethistic colonies still maintain a small population of explorers.  The dip in cooperative population observed near the end of the simulation is caused by two anomalous colonies from the simulation runs that starved to death.  No such starvations were observed among the polyethistic colonies.  We observe some population instability in this environment.  

In the roaming patch environment we see that the polyethistic colonies are able to maintain a higher population than the purely exploitative colony (the explorers quickly starve in this environment as well).  This implies a better ability to adapt to the moving patch.  The exploitative colony also displays a greater instability in population though all three successful colonies have greater instability (than in the stationary patch environment).  Also noteworthy is that all colonies have trouble maintaining an optimal population (even though the polyethistic colonies occasionally reach 80 workers).

In the seasonal environment we again observe better performance from the polyethistic colonies than the purely explorer and purely exploiter colonies.  Further there is greater stability of population in the polyethistic colonies, where the pure explorer and pure exploiter colonies suffer population oscillations corresponding roughly to the changing seasons.  Note in the figures the dotted lines display the changing seasons.  The polyethistic colonies manage to maintain roughly optimal populations in this environment while the explorer colony suffers the most in the seasons when food becomes isolated in a patch.  

Finally, in the mixed environment, we again see a population advantage to polyethism.  While both the purely explorer and purely exploiter colonies survive in the mixed environment they are unable to reach the optimal populations and display a slightly greater instability.  The purely explorer population also maintains a slight population advantage over the purely exploiter population.

A secondary focus of our simulations was on the division of labor in the polyethistic colonies.  We gathered data on how many workers of each type were deployed at a given time by the polyethistic colonies.  This data is presented in Figure 3 (right column) for each environment.  We display only the number of explorer workers in the chart in contrast to the total worker population, with the number of exploitative workers being the difference.  In the caste polyethism colonies this corresponded to how many workers of each caste were available.  In the age polyethism colonies this corresponded to how many workers were currently assigned to each task, exploring or exploiting.

In the control environments the polyethistic colonies stabilized around a constant number of explorers.  For the uniform environment both colonies settled at just over half of the workers (about 50 out of 80 workers) dedicated to exploring.  It is worth noting that the colonies did not try to maximize the number of explorers in this environment.  In the patch environment the caste colony settled at around 5 workers dedicated to exploring while the age colony maintained a slightly higher number of explorer, typically oscillating between 5 and 15 workers.  We note that in these environments the age polyethistic colony displayed greater oscillations of worker assignments whereas the caste polyethistic colony tended to stabilize around a particular division of workers assigned to each task.

In the roaming patch environment more explorers were maintained than in the stationary patch environment.  In the caste colony just over 10 of the workers were assigned the exploring role.  The age colony still assigned more workers to exploring on average than the caste colony, typically above 15, but as high as 25.  Again the age colony had greater variation in its division of labor.

The seasonal environment sees distinct performance differences among the two colonies displaying polyethism.  The caste colony settles on 30 to 35 workers dedicated to exploring.  This number is stable when compared to the age colonies that attempted to adjust the worker base to the current season.  Thus we see the number of explorers oscillating between about 25 workers to as high as 43 workers (excepting the early spike).

In the mixed environment both polyethistic colonies stabilize their worker base by assigning roughly half the workers to each task.  The age colony again assigns slightly more workers to exploration than the caste colony and displays slight oscillations.     

\section{Discussion}

The data presented suggests that polyethism, regardless of kind, offers benefits to the foraging task.  While both of the foraging methods studied in this experiment (exploring and exploiting) can be seen as self-organizing methods, the colonies benefit if the ``higher-level'' self-organizing method of polyethism is applied to select which of the methods to engage in (\cite{G10}).

In the control environments where the environment is specifically created to favor one of the two basic strategies, exploring or exploiting, we see that polyethism allows the colony to adjust the worker base to the environment.  Whereas the non-polyethistic colonies perform sub-optimally when matched with an environment they are not specialized for, the polyethistic colonies can modify their behavior to either of the environments and perform near optimally.  The only drawback in these environments to the polyethistic colonies is that they require some time to adjust to the environment.

In the more dynamic environments we see that polyethism is necessary to get optimal or near-optimal performance.  For instance we see that in the roaming patch environment, while exploiters are designed for this environment, maintaining a small population of explorers allows the new patch location to be found quicker, and more quickly exploited.  The instability seen in these environments is likely due to the shifting location of the food patch, and since the new patches are placed randomly (independent from the old patch location) each switch imposes a different level of difficulty upon the colony.  If the new patch is located close to the old one it might be more easily found and exploited than one that is far away.  However, closer patches might be exploited by extending the old path system to the new patch instead of forming a new, shorter path.  These dynamics affect foraging time in the short and long term and thus we see greater instability in population.

It is clear in the seasonal and mixed environment that polyethism is necessary to have optimal foraging.  In the seasonal environment the non-polyethistic colonies suffer in seasons where they are not favored.  In the mixed environment the non-polyethistic colonies are unable to exploit all the food drops and thus cannot maintain as high a population.  In the mixed environment the polyethistic colonies settle on a division of workers among the two strategies that allows for exploiting both food sources.  It is interesting to note that the polyethistic colonies still managed to reach optimal population levels in the mixed environment, implying that the two strategies did not experience negative interference.

\begin{figure*}
\begin{center}
\includegraphics{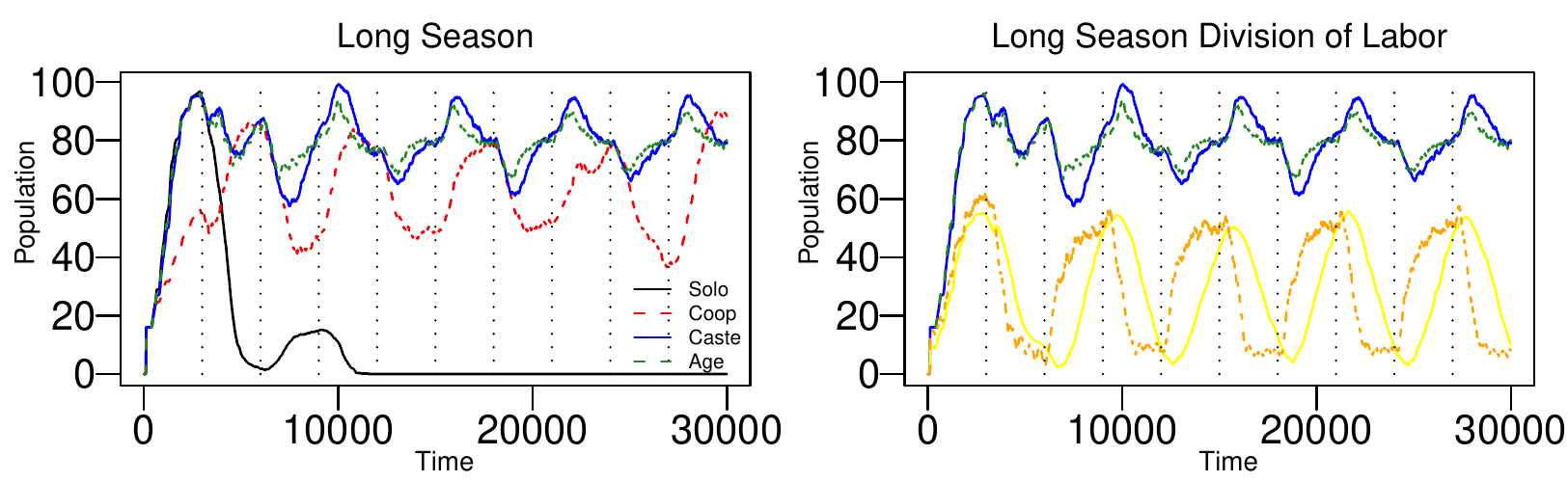}
\caption{Long (3000 round) Season.  The left chart displays population over time for the four colony types.  The right chart displays the division of labor in the Caste and Age colonies by contrasting the total population to the number of workers in the exploration task (the legend has been removed for clarity though we follow the same format as Figure 1).  Dotted lines indicate the changing seasons.}
\end{center}
\end{figure*}

We also note that the polyethistic colonies tackle the seasonal environment in slightly different ways.  In the seasonal environment the caste colony maintains a constant number of each worker.  This can be seen as the colony being prepared for either season, but not necessarily specializing for the current season.  This approach may be favored by the caste colony because the season length (1000) is short compared to the lifespan of a worker (selected uniformly from the range 2750-3250 rounds).  Thus the caste colony will not have the opportunity to adjust the balance of workers each season since workers from the previous season will still be present in the work force.  The age colony does adjust its work force to the new season, albeit only slightly, since the workers in this colony can switch tasks every round trip which is about 300-400 rounds long, shorter than the season length.  Both strategies allow the colony to maintain fairly stable and nearly optimal populations.

To test this analysis we conducted a follow up experiment where the season size was extended to 3000 rounds (see Figure 4).  In this run we saw that the caste colony adopted the adjustment strategy as well, attempting to match workers to the season instead of opting for an equal distribution.  We observed in this case that the age colony was able to adapt its workers more rapidly than the caste colony, and thus had a slightly more stable population.  The stability of both colonies' populations suffered with the longer seasons due to more polarization of the workforce and the lag between the season change and the ability of the colony to adjust its workforce.

The results of our experiment also have applications in machine learning in dealing with the exploration/exploitation trade-off.  The colonies engaged in polyethism are able to organize their foraging strategy around exploration or exploitation based on simple to form estimates of the utility of these methods.  We believe that the methods displayed by these colonies can be easily adapted to machine learning applications and have similarities to some machine learning strategies for tackling the exploration/exploitation trade-off. 

\section{Conclusion}

We conclude that division of labor is beneficial to ant colonies in that it adds a layer of dynamism to their problem solving as well as makes the colony more robust.  We suggest that the simple self-organizing methods of assigning workers to tasks can be adopted in artificial systems.  These methods are simple to implement and require a minimal amount of central planning or control.  The methods are reactive and dynamic and can likely be applied in a variety of situations, this being the topic of future work.

While we found little evidence favoring one of age or caste polyethism as a method of assigning workers to tasks we did find that the caste polyethism appeared to be more rigid in that it took longer for the workforce to adjust to new conditions.  However, the trade off is that in the age polyethistic colonies there was a tendency to over adjust to new conditions, which may not be favorable in all situations.  We believe that more work is required to determine the benefits of each of these methods, given that the distribution of these methods among natural colonies is not balanced (recall age polyethism is more common than caste polyethism).  One aspect that was not considered in this experiment, and probably plays an important role in natural colonies, is the variable costs to a colony or species (genetically and in terms of energy expenditure) in producing workers that either are specialized for their task (caste polyethism) or are generalists able to take on any available task (age polyethism).

\footnotesize
\bibliography{polyethism}
\bibliographystyle{apalike}
\end{document}